\documentclass{article}

\usepackage[preprint,nonatbib]{neurips_2020}
\usepackage[utf8]{inputenc} 
\usepackage[T1]{fontenc}    
\usepackage{hyperref}       
\usepackage{url}            
\usepackage{booktabs}       
\usepackage{amsfonts}       
\usepackage{nicefrac}       
\usepackage{microtype}      
\usepackage{amssymb,amsthm,mathtools,color,subfig}
\usepackage[ruled]{algorithm2e}
\title{Task-similarity Aware Meta-learning through Nonparametric Kernel Regression}
 \author{%
	Arun Venkitaraman\\
	Division of Decision and Control Systems\\
	KTH Royal Institute of Technology\\
	Stockholm, Sweden \\
	\texttt{arunv@kth.se} \\
		\And Anders Hansson\\
	Department of Electrical Enginerring\\
	Link\"opings University \\
    Sweden \\
	\texttt{anders.g.hansson@liu.se} \\
	\And Bo Wahlberg \\
	Division of Decision and Control Systems\\
	KTH Royal Institute of Technology \\
	Stockholm, Sweden \\
	\texttt{bo@kth.se} \\	
	}
\begin{document}
	
	\maketitle
	
	\begin{abstract}
	
This paper investigates the use of nonparametric kernel-regression to obtain a task- similarity aware meta-learning algorithm. Our hypothesis is that the use of task-similarity helps meta-learning when the available tasks are limited and may contain outlier/ dissimilar tasks. While existing meta-learning approaches implicitly assume the tasks as being similar, it is generally unclear how this task-similarity could be quantified and used in the learning. As a result, most popular meta-learning approaches do not actively use the similarity/dissimilarity between the tasks, but rely on availability of huge number of tasks for their working. Our contribution is a novel framework for meta-learning that explicitly uses task-similarity in the form of kernels and an associated meta-learning algorithm. We model the task-specific parameters to belong to a reproducing kernel Hilbert space where the kernel function captures the similarity across tasks. The proposed algorithm iteratively learns a meta-parameter which is used to assign a task-specific descriptor for every task. The task descriptors are then used to quantify the task-similarity through the kernel function. We show how our approach conceptually generalizes the popular meta-learning approaches of model-agnostic meta-learning (MAML) and Meta-stochastic gradient descent (Meta-SGD) approaches. Numerical experiments with regression tasks show that our algorithm outperforms these approaches when the number of tasks is limited, even in the presence of outlier or dissimilar tasks. This supports our hypothesis that task-similarity helps improve the meta-learning performance in task-limited and adverse settings.
	\end{abstract}
	
	\section{Introduction}
	Meta-learning seeks to abstract a general learning rule applicable to a class of different learning problems or tasks, given the knowledge of a set of training tasks from the class \cite{DBLP:conf/iclr/FinnL18,DBLP:conf/nips/DeneviCSP18}. 
	The setting is such that the data available for solving each task is often severely limited, resulting in a poor performance when the tasks are solved individually. This also sets meta-learning apart from the transfer learning paradigm where the focus is to transfer a well-trained network from existing domain to another \cite{5288526}.
%
%
{\color{black}	
 While existing meta-learning approaches implicitly
assume the tasks as being similar, it is generally unclear how this task-similarity
could be quantified and used in the learning. As a result, most popular meta-learning approaches do not actively use the similarity/dissimilarity between the
tasks, but rely on availability of huge number of tasks for their working.
	In many practical applications, the number of tasks could be limited and the tasks may not always be very similar. There might even be `outlier tasks' or `out-of-the-distribution tasks' that are less similar or dissimilar from the rest of the tasks. Our conjecture is that {\em the explicit incorporation or awareness of task-similarity helps improve meta-learning performance in such task-limited and adverse settings}. 
	
	The goal of this paper is to test this hypothesis by developing a task-similarity aware meta-learning algorithm using nonparametric kernel regression.} Specifically, our contribution is a novel meta-learning algorithm called the {\it Task-similarity Aware Nonparametric Meta-Learning} (TANML) that:
 \begin{itemize}
 \setlength\itemsep{-0.25em}
     \item Explicitly {\it employs similarity} across the tasks to fast adapt the meta-information to a given task, by using nonparametric kernel regression. 
      \item Models the parameters of a task as belonging to a reproducing kernel Hilbert space (RKHS), obtained by viewing the popular meta-learning of MAML and Meta-SGD approaches through the lens of linear/kernel regression.
      \item Assigns a {\em task-descriptor} to every task that is used to quantify task-similarity/dissimilarity through a kernel function.
     \item Offers a {\it general framework} for incorporating task-similarity in the meta-learning process. Though we pursued the algorithm with a specific choice of the task-descriptors, the proposed RKHS task-similarity aware framework can be extended to other formulations.
 \end{itemize}
 {\color{black} We  wish to emphasize that our goal is {\em not} to propose another meta-learning algorithm that outperforms the state-of-the-art, but rather to investigate if task-similarity can be explicitly incorporated and used to advantage in a meaningful manner. We show how this is achieved as the consequence of viewing the popular MAML and Meta-SGD formulations through the lens of nonparametric kernel regression. In order to keep the comparison fair on an apple-to-apple level, we compare the performance of TANML with that of MAML and Meta-SGD algorithms.}

{\color{black}
\subsection{Mathematical overview of the proposed task-similarity aware framework}
Given pairs of data $(x_{\mathrm{k}},y_{\mathrm{k}})\in\mathbb{R}^{n_x}\times\mathbb{R}^{n_y}$ where ${\mathrm{k}}\in\{1,2,\cdots,K\}$ {\color{black}generated by an unknown data source}, we are interested in learning a predictor $\mathbb{R}^{n_x}\times\mathbb{R}^D\ni(x,\pmb\theta)\mapsto f(x,\pmb\theta)\in\mathbb{R}^{n_y}$ from the given data. For example, $ f(x,\pmb\theta)$ could be {\color{black} a} function {\color{black}defined} by an artificial neural network {\color{black}(ANN)}. We collect pairs of data in $\mathcal{X}=(x_1,x_2,\cdots,x_K)$ and $\mathcal{Y}=(y_1,y_2,\cdots,y_K)$ and define the loss function $\mathbb{R}^{Kn_x}\times \mathbb{R}^{Kn_y}\times\mathbb{R}^{D}\ni(\mathcal{X},\mathcal{Y},\pmb\theta)\mapsto\mathcal{L}(\mathcal{X},\mathcal{Y},\pmb\theta)\in\mathbb{R}$ which we then minimize with respect to $\pmb\theta$. This constitutes the training of the predictor. In the case of a ANN, $\mathcal{L}(\mathcal{X},\mathcal{Y},\pmb\theta)\in\mathbb{R}$ could be the mean-square loss or the cross-entropy function. The data $\mathcal{X},\mathcal{Y}$ used for training is referred to as the training data. Let $\hat{\pmb\theta}$ denote the optimal value of $\pmb\theta$ obtained from training. Given a new $x\in\mathbb{R}^{n_x}$, we use $\hat{y}=f(x,\hat{\pmb\theta})$ to predict $y$. The goodness of $\hat{\pmb\theta}$ is evaluated using $y-\hat{y}$ on a sequence of pairs of new data called the test data $\bar{\mathcal{X}}$, $\bar{\mathcal{Y}}$, defined similarly as $\mathcal{X}$ and $\mathcal{Y}$, but with $\bar{K}$ number of data pairs. {\color{black} The training of the predictor for the given data source is defined as a task}. 

{\color{black} Now, consider that we are interested in carrying out several such tasks for data coming from different but similar sources.} Let  $\mathcal{X}_i,\mathcal{Y}_i,\bar{\mathcal{X}}_i,\bar{\mathcal{Y}}_i,\,i=1,\cdots,T_{tr}$ denote the data from $T_{tr}$ different data-sources, and defined similarly as $\mathcal{X},\mathcal{Y},\bar{\mathcal{X}},\bar{\mathcal{Y}}$ above. {\color{black} We refer to the training of the predictor for data $\mathcal{X}_i,\mathcal{Y}_i,\bar{\mathcal{X}}_i,\bar{\mathcal{Y}}_i$ as the $i$th training task, and $\pmb\theta_i$ is referred to as the parameter for the task.} Meta-learning captures similarities across the tasks by learning a common $\hat{\pmb\theta}$ (which we denote by $\pmb\theta_0$) from the data of these $T_{tr}$ tasks (called the meta-training data), such that $\pmb\theta_0$ can be quickly adapted to train a predictor for data from {\color{black}new and different but similar data-sources.} Depending on how $\pmb\theta$ is obtained from $\pmb\theta_0$, various meta-learning algorithms exist \cite{DBLP:conf/nips/DeneviCSP18,DBLP:conf/iclr/FinnL18,DBLP:conf/icml/AllenSST19}. The performance of the meta-learning algorithm is evaluated on previously unseen data from several other similar sources $\mathcal{X}^v_i,\mathcal{Y}^v_i,\bar{\mathcal{X}}^v_i,\bar{\mathcal{Y}}^v_i,\,i=1,\cdots,T_{v}$ (called the meta-test data) defined similarly to $\mathcal{X},\mathcal{Y},\bar{\mathcal{X}},\bar{\mathcal{Y}}$ $-$ this constitutes the meta-test phase. {\color{black} The training of the predictor for test data $\mathcal{X}^v_i,\mathcal{Y}^v_i,\bar{\mathcal{X}^v}_i,\bar{\mathcal{Y}^v}_i$ is referred to as the $i$th test task, $\pmb\theta^v_i$ denotes the parameter for the $i$th test task.} In the existing meta-learning approaches, both ${\pmb\theta}_i$ and $\pmb\theta^v_i$ are obtained by adapting $\pmb\theta_0$ using the gradient  of $\mathcal{L}(\mathcal{X}_i,\mathcal{Y}_i,\pmb\theta)$ and $\mathcal{L}(\mathcal{X}^v_i,\mathcal{Y}^v_i,\pmb\theta)$, respectively, evaluated at $\pmb\theta_0$.

{\color{black}In our work, we propose a meta-learning algorithm where ${\pmb\theta}_i$ explicitly uses a similarity between the $i$th training task and all the training tasks. Similarly, the parameters $\pmb\theta^v_i$ for the test tasks also use explicitly a similarity between the $i$th test task and all the training tasks. As specified later, we define this {\em task-similarity} between two tasks through kernel regression, and our algorithm learns the kernel regression coefficients $\pmb\Psi$ as meta-parameters in addition to $\pmb\theta_0$.} 
%
%
%
\label{sec:intro}
}
\paragraph{A motivating example}
{\color{black}Let us now consider a specific loss function given by}
$
\mathcal{L}(\mathcal{X},\mathcal{Y},\pmb\theta)=\sum_{{\mathrm{k}}=1}^K\|y_{\mathrm{k}}-f(x_{\mathrm{k}},\pmb\theta)\|_2^2.
$
 Training for tasks individually will result in a {\color{black}predictor} that overfits to $\mathcal{X},\mathcal{Y}$, and generalizes poorly to $\bar{\mathcal{X}},\bar{\mathcal{Y}}$. MAML-type meta-learning approaches \cite{DBLP:conf/icml/FinnAL17} solve this by inferring the information across tasks in the form of a good initialization $\pmb\theta_0$-- specialized/adapted to a new task {\color{black} using the {\em adaptation function} $\mathbb{R}^{D}\times\mathbb{R}^{Kn_x}\times \mathbb{R}^{Kn_y}\ni(\pmb\theta_0,\mathcal{X},\mathcal{Y})\mapsto g_{\mathrm{MAML}}(\pmb\theta_0,\mathcal{X},\mathcal{Y})\in\mathbb{R}^D$ defined as:
 \begin{equation*}
     g_{\mathrm{MAML}}(\pmb\theta_0,\mathcal{X},\mathcal{Y})\triangleq\pmb\theta_0-\alpha\nabla_{\pmb\theta_0}\mathcal{L}(\mathcal{X},\mathcal{Y},\pmb\theta_0)
 \end{equation*}
The parameters for the training and test tasks as obtained through adaptation of $\pmb\theta_0$ as
\begin{align*}
    &\pmb\theta_i=g_{\mathrm{MAML}}(\pmb\theta_0,\mathcal{X}_i,\mathcal{Y}_i),\,\, i=1,\cdots,T_{tr},\,\,\mbox{and }\,
    &\pmb\theta^v_i=g_{\mathrm{MAML}}(\pmb\theta_0,\mathcal{X}^v_i,\mathcal{Y}^v_i),\,\, i=1,\cdots,T_v.
\end{align*}}
The meta-parameter $\pmb\theta_0$ is learnt by iteratively taking a gradient descent with respect to the test loss on the training tasks given by {\color{black}$\sum_{i=1}^{{T}_{tr}}\mathcal{L}(\bar{\mathcal{X}}_i,\bar{\mathcal{Y}}_i,g_{\mathrm{MAML}}(\pmb\theta_0,\mathcal{X}_i,\mathcal{Y}_i))$.}
{\color{black} The parameters for a task are obtained directly from $\pmb\theta_0$ and does not make use of any information from the other training tasks. As a result, the common $\pmb\theta_0$ learnt during the meta-training treats all tasks equally $-$the algorithm implicitly assumes similarity of all tasks, but is not able to discern or quantify the degree of similarity or dissimilarity among the tasks.

In contrast, our algorithm involves an adaptation function $g_{\mathrm{TANML}}$ (to be defined later) that explicitly uses a notion of similarity between the tasks to predict parameters for a task. As a result, we expect that our approach helps train predictors even when the data-sources that are not very similar to each other. In our numerical experiments in Section 4, we see that this is indeed the case the sinusoidal function as the data source. }
	\subsection{Related work}
    The structural characterization of tasks and use of task-dependent knowledge has gained interest in meta-learning recently. 
	In \cite{DBLP:conf/iclr/EdwardsS17}, a variational autoencoder based approach was employed to generate task/dataset statistics  used to measure similarity. In \cite{DBLP:conf/emnlp/RuderP17}, domain similarity and diversity measures were considered in the context of  transfer learning \cite{DBLP:conf/emnlp/RuderP17}. The study of how task properties affect the catastrophic forgetting in continual learning was pursued in \cite{DBLP:journals/corr/abs-1908-01091}. In \cite{DBLP:conf/iclr/LeeLNKPYH20}, the authors proposed a task-adaptive meta-learning approach for classification that adaptively balances meta-learning and task-specific learning differently for every task and class. It was shown in \cite{DBLP:conf/nips/OreshkinLL18} that the performance few-shot learning shows significant improvements with the use of task-dependent metrics. While the use of kernels or similarity metrics is not new in meta-learning, they are typically seen in the context of defining relations between the classes or samples within a given task \cite{DBLP:conf/nips/VinyalsBLKW16,NIPS2017_6996,DBLP:conf/nips/OreshkinLL18,DBLP:journals/corr/abs-1901-08098,goo2020local}.
	Information-theoretic ideas have also been used in the study of the topology and the geometry of task spaces \cite{DBLP:journals/corr/abs-1908-01091,aless2018dynamics}. In \cite{DBLP:conf/iccv/AchilleLTRMFSP19}, the authors construct vector representations for tasks using partially trained probe networks, based on which task-similarity metrics are developed.  Task descriptors have been of interest specially in vision related tasks in the context of transfer learning \cite{Zamir_2018_CVPR,DBLP:conf/iccv/AchilleLTRMFSP19,DBLP:conf/iccv/TranNH19}. Recently, neural tangent kernels were been proposed for asymptotic analysis of meta-learning for infinitely wide neural networks by considering gradient based kernels across tasks \cite{wang2020global}.
	%
\section{Review of MAML and Meta-SGD}
To facilitate our analysis, we first review MAML and Meta-SGD approaches and highlight the relevant aspects necessary for our discussion. We shall then show how these approaches lead to the definition of a generalized meta-SGD and consequently, to our TANML approach.
\subsection{Meta Agnostic Meta-Learning}
Model-agnostic meta-learning proceeds in two stages iteratively. {\color{black} As discussed in the motivating example, the parameter $\pmb\theta_i$ for the $i$th training task $\mathcal{X}_i,\mathcal{Y}_i,\bar{\mathcal{X}_i},\bar{\mathcal{Y}_i}, \,\,i=1,\cdots,T_{tr}$ is obtained by applying the adaptation function $g_{\mathrm{MAML}}$ to $\pmb\theta_0$ as $
\pmb\theta_i=g_{\mathrm{MAML}}(\pmb\theta_0,\mathcal{X}_i,\mathcal{Y}_i).
$
{\color{black}This is called the inner update. Once $\pmb\theta_i$ is obtained for all the training tasks, $\pmb\theta_0$ is then updated by running one gradient descent step on the total test-loss given by
$\sum_{i=1}^{{T}_{tr}}\mathcal{L}(\bar{\mathcal{X}_i},\bar{\mathcal{Y}_i},g_{\mathrm{MAML}}(\pmb\theta_0,\mathcal{X}_i,\mathcal{Y}_i)).
$
This is called the outer update. Each outer update involves $T_{tr}$ inner updates corresponding to the $T_{tr}$ training tasks.
The outer updates are run for $N_{iter}$ iterations.} This constitutes the meta-training phase of MAML, described in Algorithm \ref{algo:maml}. }
{\color{black}Once the meta-training phase is complete and $\pmb\theta_0$ is learnt, the parameters for a new test task are obtained by applying the inner update on the training data of the test task.} We note here that MAML described in Algorithm 1 is the first-order MAML \cite{DBLP:conf/nips/FinnXL18}, as opposed to the general MAML where the inner update may contain several gradient descent steps. We note that when we talk of MAML in this paper, we always refer to the first-order MAML. A schematic of MAML is presented in Figure 1.
\begin{center}\begin{algorithm}[H]
		\caption{Model agnostic meta-learning (MAML)}
	\SetAlgoLined
	Initialize $\pmb\theta_{0}$\\
	\For{$N_{iter}$ iterations} {
		\For{$i=1,\cdots, {T}_{tr}$}
		{${\color{black}g_{\mathrm{MAML}}(\pmb\theta_0,\mathcal{X}_i,\mathcal{Y}_i)}=\pmb\theta_{0}-\alpha\nabla_{\pmb\theta_0} \mathcal{L}(\mathcal{X}_i,\mathcal{Y}_i,\pmb\theta_0)$\quad [Inner update]}
		
		$\pmb\theta_{0}=\pmb\theta_{0}-\beta\nabla_{\pmb\theta_0}\sum_{i=1}^{{T}_{tr}}\mathcal{L}(\bar{\mathcal{X}_i},\bar{\mathcal{Y}_i},{\color{black}g_{\mathrm{MAML}}(\pmb\theta_0,{\mathcal{X}}_i,{\mathcal{Y}}_i)})$\quad[Outer update]
	}
	\label{algo:maml}
\end{algorithm}
\end{center}
 
%
\subsection{Meta-Stochastic Gradient Descent (Meta-SGD)}
Meta stochastic gradient descent (Meta-SGD) is a variant of MAML that learns the component-wise step sizes for the inner update jointly with $\pmb\theta_0$. Let $\pmb\alpha$ denote the vector of step-sizes for the different components of $\pmb\theta$. As with MAML, the meta-training phase of Meta-SGD also involves an inner and an outer update. {\color{black}The outer update computes the values of $\pmb\theta_0$ and $\pmb\alpha$; the inner update computes the parameter values $\pmb\theta_i$ using the adaptation function $\mathbb{R}^{D}\times\mathbb{R}^{D}\times\mathbb{R}^{Kn_x}\times \mathbb{R}^{Kn_y}\ni(\pmb\theta_0,\pmb\alpha,\mathcal{X}_i,\mathcal{Y}_i)\mapsto g_{\mathrm{MSGD}}(\pmb\theta_0,\pmb\alpha,\mathcal{X}_i,\mathcal{Y}_i)\in\mathbb{R}^D$ defined as
\begin{equation*}
    g_{\mathrm{MSGD}}(\pmb\theta_0,\pmb\alpha,\mathcal{X}_i,\mathcal{Y}_i)\triangleq\pmb\theta_0-\pmb\alpha\cdot\nabla_{\pmb\theta_0} \mathcal{L}(\mathcal{X}_i,\mathcal{Y}_i,\pmb\theta_0),
\end{equation*}
where $\cdot$ operator denotes the point-wise vector product. The outer update is run for $N_{iter}$ iterations.}
The meta-training phase for Meta-SGD is described in Algorithm \ref{algo:metasgd}:
\begin{algorithm}[H]
		\SetAlgoLined
		Initialize $[\pmb\theta_{0},\pmb\alpha]$\\
		\For{$N_{iter}$ iterations} {
			\For{$i=1,\cdots,{T}_{tr}$}
			{${\color{black}g_{\mathrm{MSGD}}(\pmb\theta_0,\pmb\alpha,\mathcal{X}_i,\mathcal{Y}_i)}=\pmb\theta_{0}-\pmb\alpha\cdot\nabla_{\pmb\theta_0} \mathcal{L}(\mathcal{X}_i,\mathcal{Y}_i,\pmb\theta_0)$\quad [Inner update]}
			
			$[\pmb\theta_{0},\pmb\alpha]=[\pmb\theta_{0},\pmb\alpha]-\beta\nabla_{[\pmb\theta_{0},\pmb\alpha]}\sum_{i=1}^{{T}_{tr}}\mathcal{L}(\bar{\mathcal{X}_i},\bar{\mathcal{Y}_i},{\color{black}g_{\mathrm{MSGD}}(\pmb\theta_0,\pmb\alpha,{\mathcal{X}}_i,{\mathcal{Y}}_i))}$\quad[Outer update]
		}
		\label{algo:metasgd}
	\caption{Meta-stochastic gradient descent}
	\end{algorithm}
The predictor for the $i$th test task is then trained by applying the inner update on $\mathcal{X}^v_i,\mathcal{Y}^v_i$, using the values of $\pmb\theta_0$ and $\pmb\alpha$ obtained in the meta-training phase.

We notice that the inner update is expressible as $ {\color{black}g_{\mathrm{\tiny{MSGD}}}(\pmb\theta_0,\pmb\alpha,\mathcal{X}_i,\mathcal{Y}_i)}=\mathbf{W}^\top\mathbf{z}_i(\pmb\theta_0)$ where
\begin{eqnarray}
\label{eq:metasgd2}
\mathbf{W}\triangleq\left[\begin{matrix}
\mathbf{I}, 
-\mbox{diag}(\pmb\alpha)
\end{matrix}\right]\,\,\quad \mbox{and}\quad\,\, \mathbf{z}_i(\pmb\theta_0)\triangleq\left[\begin{matrix}
\pmb\theta_0^\top\,
\nabla_{\pmb\theta_0},\, \mathcal{L}(\mathcal{X}_i,\mathcal{Y}_i,\pmb\theta_0)^\top
\end{matrix}\right]^\top.
\end{eqnarray}
The matrix $\mathbf{W}^\top$ denotes the transpose of $\mathbf{W}$, $\mathbf{I}$ denotes the identity matrix, and $\mbox{diag}(\pmb\alpha)$ denotes the diagonal matrix whose diagonal is equal to the vector $\pmb\alpha$. We refer to $\mathbf{z}_i(\pmb\theta_0)$ as the {\it task descriptor} of the $i$th training task. 
{\color{black}Thus, $g_{\mathrm{MSGD}}(\pmb\theta_0,\pmb\alpha,\mathcal{X}_i,\mathcal{Y}_i)$ takes the form of a linear predictor for $\pmb\theta_i$ with $\mathbf{z}_i(\pmb\theta_0)$ as the input and regression coefficients $\mathbf{W}$. The adaptation $g_{\mathrm{MSGD}}(\pmb\theta_0,\pmb\alpha,\mathcal{X}_i,\mathcal{Y}_i)$ can be generalized to the case of $\mathbf{W}$ being a full matrix that is to be learnt from the training tasks.
This generalization results in the adaptation function 
$\mathbb{R}^{D}\times\mathbb{R}^{D\times 2D}\times\mathbb{R}^{Kn_x}\times \mathbb{R}^{Kn_y}\ni(\pmb\theta_0,\mathbf{W},\mathcal{X},\mathcal{Y})\mapsto g_{\mathrm{GMSGD}}(\pmb\theta_0,\mathbf{W},\mathcal{X},\mathcal{Y})\in\mathbb{R}^D$ given by
\begin{equation*}
{\color{black}g_{\mathrm{GMSGD}}(\pmb\theta_0,\mathbf{W},\mathcal{X}_i,\mathcal{Y}_i)}=\mathbf{W}^\top\mathbf{z}_i(\pmb\theta_0)=\mathbf{W}^\top_1\pmb\theta_0+\mathbf{W}^\top_2\nabla_{\pmb\theta_0} \mathcal{L}(\mathcal{X}_i,\mathcal{Y}_i,\pmb\theta_0)
\end{equation*}
where $\mathbf{W}_1,\mathbf{W}_2\in\mathbb{R}^{D\times D}$ are the submatrices of $\mathbf{W}$ such that $\mathbf{W}=[\mathbf{W}_1\, \mathbf{W}_2]$. {\color{black}Expressed in this manner, we notice how $g_{\mathrm{GMSGD}}$ performs a parameter update similar to a second-order gradient update with $\mathbf{W}_2$ taking a role similar to the Hessian matrix.}
We refer to the resulting meta-learning algorithm as the {\em Generalized Meta-SGD} described in Algorithm \ref{algo:gensgd}}. The second term $\Omega(\mathbf{W})$ in the outer loop cost function is a regularization that ensures $\mathbf{W}$ is bounded and avoids overfitting.
{\color{black} On setting $\mu=0$ and using  $\mathbf{W}$ as defined in \eqref{eq:metasgd2}, the Generalized Meta-SGD reduces to the Meta-SGD.}
   The Generalized Meta-SGD is thus a more general form of MAML arrived at by viewing MAML/Meta-SGD as a linear regression. 
	\begin{algorithm}[H]
		\SetAlgoLined
		Initialize $[\pmb\theta_{0},\mathbf{W}\in\mathbb{R}^{2D\times D}]$\\
		\For{$N_{iter}$ iterations} {
			\For{$i=1,\cdots,{T}_{tr}$}
			{${\color{black}g_{\mathrm{GMSGD}}(\pmb\theta_0,\mathbf{W},\mathcal{X}_i,\mathcal{Y}_i)}=\mathbf{W}^\top\mathbf{z}_i(\pmb\theta_0)$\quad [Inner update]}
			
			$[\pmb\theta_{0},\mathbf{W}]=[\pmb\theta_{0},\mathbf{W}]-\beta\nabla_{[\pmb\theta_{0},\mathbf{W}]}\left(\sum_{i=1}^{{T}_{tr}}\mathcal{L}(\bar{\mathcal{X}_i},\bar{\mathcal{Y}_i},{\color{black}g_{\mathrm{GMSGD}}(\pmb\theta_0,\mathbf{W},{\mathcal{X}}_i,{\mathcal{Y}}_i)})+\mu\Omega(\mathbf{W})\right)$
		}
	\caption{Generalized Meta-SGD}
	\label{algo:gensgd}
	\end{algorithm}
%
\section{Task-similarity Aware Meta-Learning}
 It is well known that the expressive power of linear regression is limited due to both its linear nature and the finite dimension of the input. Further, since the dimension of linear regression matrix $\mathbf{W}$ grows quadratically with the dimension of $\pmb\theta$, a large amount of training data would be necessary to estimate it. A transformation of linear regression in the form of {\it'kernel substitution'} or {\it 'kernel trick'} results in the more general nonparametric or kernel regression \cite{Bishop,DBLP:books/lib/ScholkopfS02}. Kernel regression essentially performs linear regression in an infinite dimensional space making it a simple yet powerful and effective nonlinear approach. {\color{black} This motivates us to use kernel regression model as the natural next step from the Generalized Meta-SGD developed in the earlier section. By generalizing the linear regression model in $g_{\mathrm{GMSGD}}$, we propose an adaptation function $\mathbb{R}^{D}\times \mathbb{R}^{D\times T_{tr}}\times\mathbb{R}^{Kn_x}\times \mathbb{R}^{Kn_y}\ni(\pmb\theta_0,\pmb\Psi,\mathcal{X},\mathcal{Y})\mapsto g_{\mathrm{TANML}}(\pmb\theta_0,\pmb\Psi,\mathcal{X},\mathcal{Y})\in\mathbb{R}^D$ in the form of nonparametric or kernel regression model given by}
\begin{equation}
\label{eq:krmaml_inner}
g_{\mathrm{TANML}}(\pmb\theta_0,\pmb\Psi,\mathcal{X}_i,\mathcal{Y}_i)=\sum_{i=1}^{{T}_{tr}}{\pmb\psi}_{i}k(\mathbf{z}_i(\pmb\theta_0),\mathbf{z}_{i}(\pmb\theta_0))={\pmb\Psi}^\top\mathbf{k}(\pmb\theta_{0},i),
\end{equation}
where ${k}:\mathbb{R}^{2D}\times\mathbb{R}^{2D}\mapsto\mathbb{R}$ denotes a valid kernel function\footnote{A valid kernel function is one that results in the kernel matrix evaluated for any number of datapoints to be symmetric and positive-semidefinite cf. \cite{Bishop}}, {\color{black}$\mathbf{k}(\pmb\theta_{0},i)\triangleq\left[\,k\left(\mathbf{z}_i(\pmb\theta_0),\mathbf{z}_{1}(\pmb\theta_0)\right), \cdots, k\left(\mathbf{z}_i(\pmb\theta_0),\mathbf{z}_{T_{tr}}(\pmb\theta_0)\right)\,\right]^\top$ is the vector with kernel values between the $i$th training task and all the $T_{tr}$ training tasks, and $\pmb\Psi=\left[\pmb\psi_1,\cdots,\pmb\psi_{T_{tr}}\right]$ is the matrix of kernel regression coefficients stacked along the columns. The kernel coefficient matrix $\pmb\Psi$ and the parameter $\pmb\theta_0$ are learnt in the meta-training phase by iteratively performing an outer update as in the case of the Generalized Meta-SGD. The computed $\pmb\Psi$ and $\pmb\theta_0$ are then used to train the predictor for a new test task by applying the inner update on its training data.}
We call our approach {\em Task-similarity Aware Nonparametric Meta-Learning (TANML)} since the kernel measures the similarity between tasks through the task-descriptors.

The kernel regression in \eqref{eq:krmaml_inner} models ${\pmb\theta}_i$ 
as belonging to the space of functions 
defined as
$\mathcal{H}$:
\begin{equation}
    \mathcal{H}=\left\{\tilde{\pmb\theta}: \tilde{\pmb\theta}=\sum_{i'=1}^{{T}_{tr}}\tilde{\pmb\psi}_{i}k(\mathbf{z}_i(\pmb\theta_0),\mathbf{z}_{i'}(\pmb\theta_0)),\,\,\, \tilde{\pmb\psi}_{i'}\in\mathbb{R}^D,\,\, i'=1,\cdots,T_{tr}\,\right\}
    \label{eq:rkhs}
\end{equation}
{\color{black}The space $\mathcal{H}$ is referred to as the {\em reproducing kernel Hilbert space} (RKHS) associated with the kernel ${k}(\cdot,\cdot)$. We refer the reader to \cite{hofmann2008} and \cite{DBLP:books/lib/ScholkopfS02} for further reading on RKHS.
The space $\mathcal{H}$ has an important structure: each function in $\mathcal{H}$ uses the information (the coefficients $\bar{\pmb\psi}$) from all the $T_{tr}$ training tasks weighted by the kernel that essentially quantifies a similarity or correlation between the tasks through the task-descriptors defined earlier. }
Computing the optimal values of the kernel coefficients $\pmb\Psi$ and $\pmb\theta_0$, which forms the meta-training phase, is then equivalent to solving the functional minimization problem:
\begin{equation*}
\arg\min_{\pmb\theta_0,\tilde{\pmb\theta}\in\mathcal{H}}\left(\sum_{i=1}^{{T}_{tr}}\mathcal{L}(\bar{\mathcal{X}_i},\bar{\mathcal{Y}_i},\tilde{\pmb\theta})+\mu\|\tilde{\pmb\theta}\|^2_\mathcal{H}\right),
   \label{eq:rkhs3}
\end{equation*}
where the regularization term is the squared-norm in the RKHS which promotes smoothness and controls overfitting, $\mu$ being the regularization constant. The squared-norm in an RKHS is defined as  $\|\tilde{\pmb\theta}\|^2_\mathcal{H}\triangleq\sum_{i=1}^{{T}_{tr}}\sum_{i'=1}^{{T}_{tr}}{\pmb\psi}_{i}{\pmb\psi}_{i'}k(\mathbf{z}_i(\pmb\theta_0),\mathbf{z}_{i'}(\pmb\theta_0))={\pmb\Psi}^\top{\color{black}\mathbf{K}(\pmb\theta_0)}{\pmb\Psi}$; and ${\color{black}\mathbf{K}(\pmb\theta_0)}\in\mathbb{R}^{T_{tr}\times T_{tr}}$ is the matrix of kernels evaluated across all the training tasks. This novel connection between meta-learning and RKHS obtained from TANML could potentially help in the mathematical understanding of existing algorithms, and help develop new meta-learning algorithms in the light of the RKHS theory \cite{hofmann2008,DBLP:books/lib/ScholkopfS02}. 

\begin{figure}
    \centering
   {\includegraphics[height=1.8in]{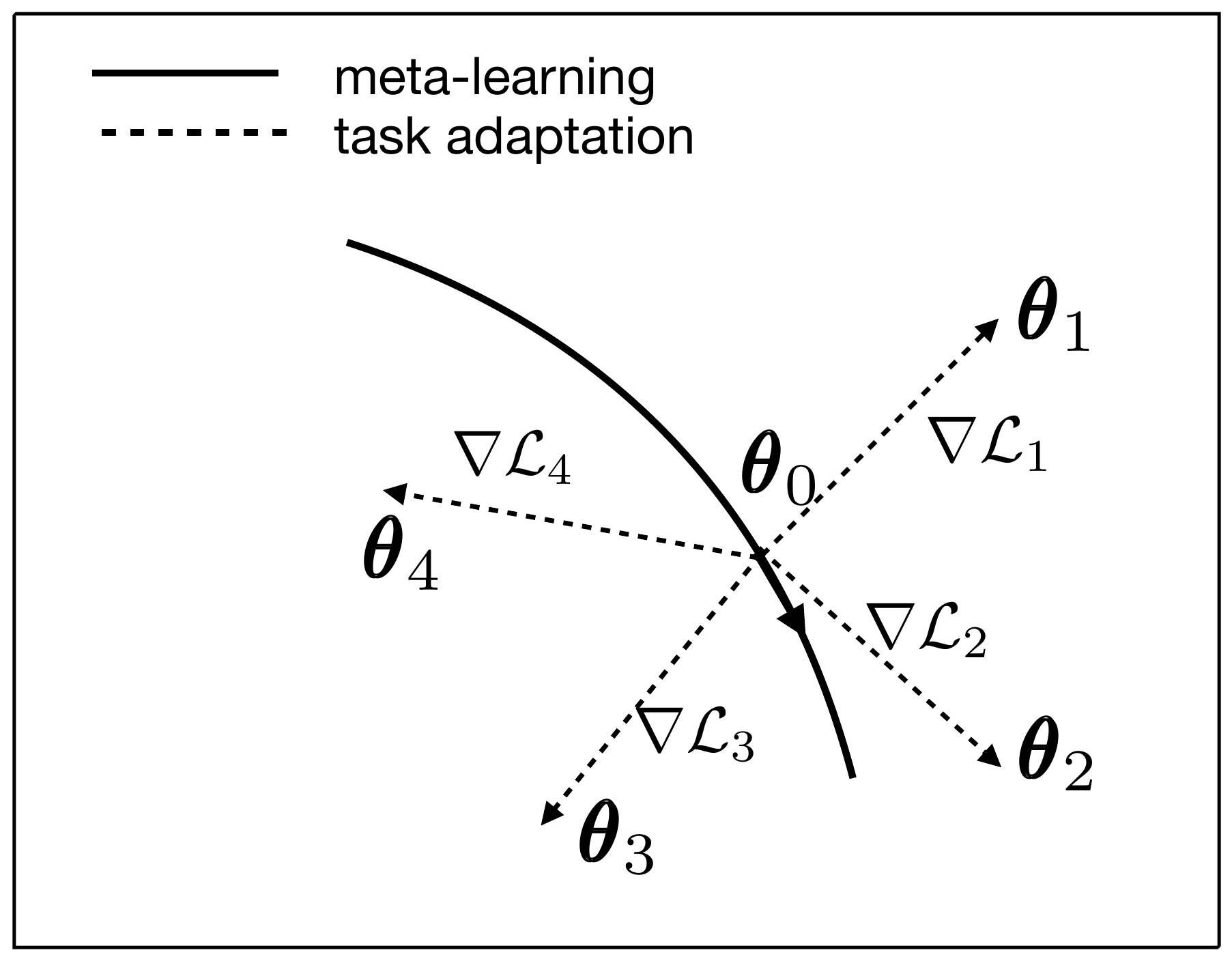}}
    {\includegraphics[height=1.8in]{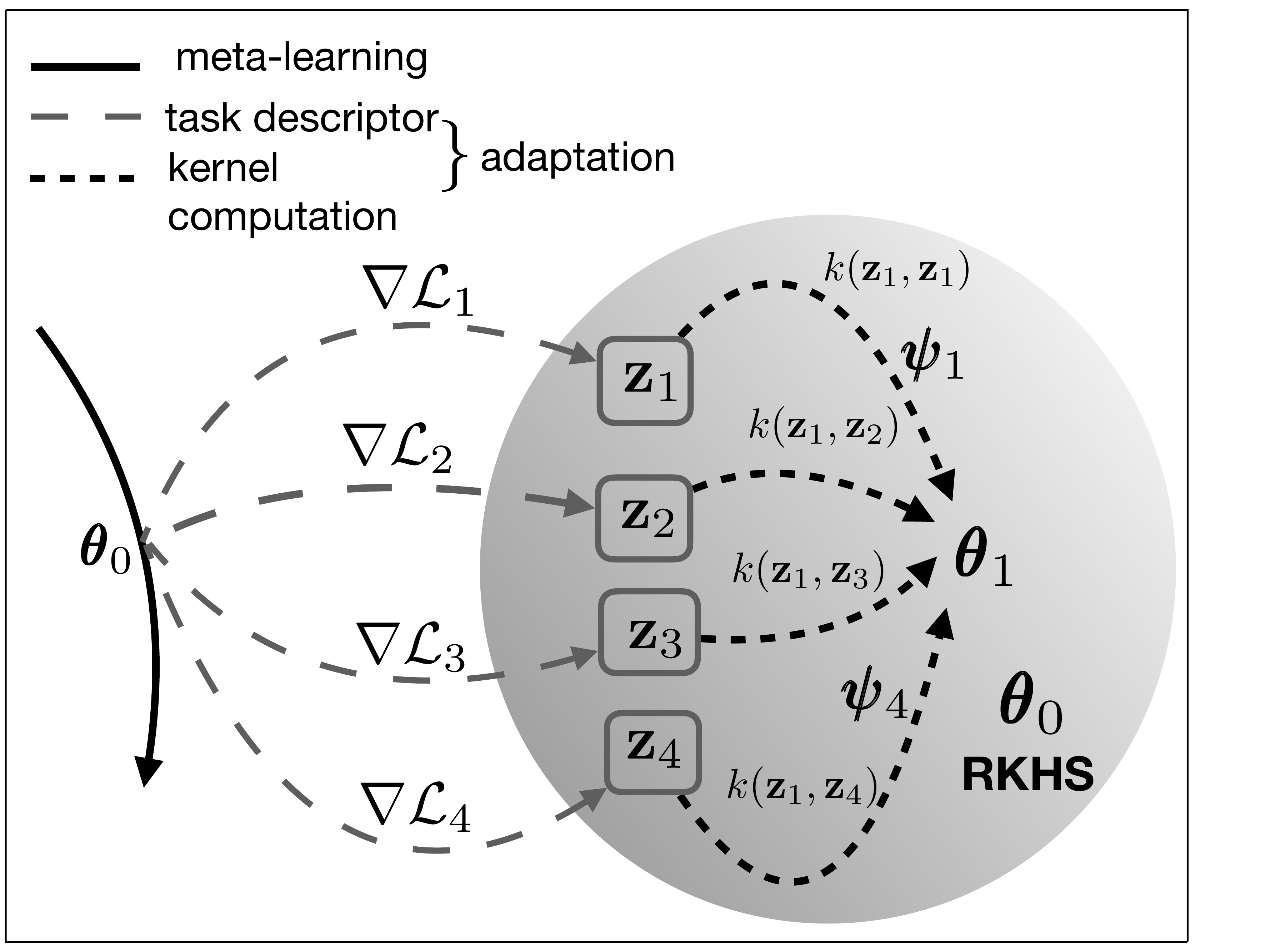}}
    \caption{{\it Left:} Schematic of MAML. 
    {\it Right: }Schematic of the TANML. Only the computation of $\pmb\theta_1$ is shown to keep the diagram uncluttered.\\
   }
    \label{fig:taml}
    \vspace{-.25in}
\end{figure}

The meta-training phase for TANML is described in Algorithm \ref{algo:TANML}, where we use $\Omega(\pmb\Psi)={\pmb\Psi}^\top{\color{black}\mathbf{K}(\pmb\theta_0)}{\pmb\Psi}$. In general, other regularizations such as $\ell_1$ or $\ell_2$ norms could also be used.
We also note from \eqref{eq:krmaml_inner} that the {\em TANML approach is a general framework}: any kernel and any task-descriptor which meaningfully captures the information in the task could be employed. What constitutes a meaningful descriptor for a task is an open question; while there have been studies on deriving features and metrics for understanding the notion of similarity between data sources or datasets \cite{DBLP:journals/corr/abs-1901-05761}, they have not been used in the actual design of meta-algorithms. The particular form of the task-descriptors used in our derivation is the result of taking MAML/Meta-SGD as a starting point, and follows naturally from analyzing them through the lens of linear and kernel regression. A schematic describing the task-descriptor based TANML and the intuition behind its working is shown in Figure \ref{fig:taml}.
	\begin{algorithm}[H]
	Initialize $[\pmb\theta_{0},\pmb\Psi\in\mathbb{R}^{T_{tr}\times D}]$\\
	\For{$N_{iter}$ iterations} {
		\For{$i=1,\cdots,{T}_{tr}$}
		{${\color{black}g_{\mathrm{\tiny TANML}}(\pmb\theta_0,\pmb\Psi,\mathcal{X}_i,\mathcal{Y}_i)}=\pmb\Psi^\top\mathbf{k}(\pmb\theta_{0},i)$\quad [Inner update]}
		
		$[\pmb\theta_{0},\pmb\Psi]=[\pmb\theta_{0},\pmb\Psi]-\beta\nabla_{[\pmb\theta_{0},\pmb\Psi]}\sum_{i=1}^{{T}_{tr}}\mathcal{L}(\bar{\mathcal{X}_i},\bar{\mathcal{Y}_i},{\color{black}g_{\mathrm{TANML}}(\pmb\theta_0,\pmb\Psi,{\mathcal{X}}_i,{\mathcal{Y}}_i)})+\mu\Omega(\pmb\Psi)$\quad[Outer update]
	}
	\caption{Task-similarity Aware Meta LearningA}
	\label{algo:TANML}
	\vspace{-.2in}
\end{algorithm}
\paragraph{On the choice of kernels and sequential training}
While the expressive power of kernels is immense, it is also known that the performance could vary depending on the choice of the kernel function\cite{DBLP:books/lib/ScholkopfS02}. The kernel function that works best for a dataset is usually found by trial and error. A possible approach is to use multi-kernel regression where one lets the data decide which of the pre-specified set of kernels are relevant \cite{multikernel_4,multikernel_7}. Domain-specific knowledge may also be incorporated in the choice of kernels. In our analysis, we use two of the popular kernel functions: the Gaussian or the radial basis function (RBF) kernel, and the cosine kernel.

We note that since MAML and Meta-SGD and similar approaches perform the inner update independently for every task, they naturally admit a sequential or batch based training. Since TANML uses an inner update in the form of a nonparametric kernel regression, it inherits one of the limitations of kernel-based  approaches $-$ that all training data is used simultaneously for every task. As a result, the task losses and the associated gradients for all the training tasks are used at every inner update of TANML. One way to overcome this limitation would be the use of online or sequential kernel regression techniques \cite{DBLP:journals/jmlr/LuHWZL16,DBLP:journals/tkdd/SahooHL19,DBLP:conf/nips/VermaakGD03}. We are currently working towards achieving this improvement to our algorithm.
%
\section{Numerical experiments}
We evaluate the performance of TAML and compare it with MAML and Meta-SGD on two synthesized regression datasets. These tasks have been used previously by previous works \cite{DBLP:conf/nips/DeneviCSP18,DBLP:conf/icml/FinnAL17,DBLP:conf/nips/FinnXL18} in meta-learning as a baseline for evaluating the performance on regression tasks. We consider two kernel functions for TANML: the Gaussian kernel $k(\mathbf{z}_{i}(\pmb\theta_0),\mathbf{z}_{i'}(\pmb\theta_0))=\exp\left(-\|\mathbf{z}_{i}(\pmb\theta_0)-\mathbf{z}_{i'}(\pmb\theta_0)\|_2^2/\sigma^2\right)$, and the cosine kernel $k(\mathbf{z}_{i}(\pmb\theta_0),\mathbf{z}_{i'}(\pmb\theta_0))=\frac{\mathbf{z}_i(\pmb\theta_0)^\top\mathbf{z}_{i'}(\pmb\theta_0)}{\|\mathbf{z}_{i}(\pmb\theta_0)\|\|\mathbf{z}_{i'}(\pmb\theta_0)\|}$. 
The performance of the various meta-learning approaches are compared using the normalized mean-squared error (NMSE) on the test tasks:
$
     \displaystyle\mbox{NMSE}\triangleq\frac{\sum_{i=1}^{T_{v}}\sum_{k=1}^K(y_k-\hat{y}_k)^2}{\sum_{i=1}^{T_{v}}\sum_{k=1}^K y_k^2}.
$
 The numerical details of the experiments not mentioned in the manuscript, such as the learning rate and other hyper-parameters, are given in the appendix for space constraints.
\subsection{Experiment 1}
{\color{black} We consider the task of training linear predictors of the form $f(x,\pmb\theta)=\pmb\theta^\top x$. The task data pairs $(x,y)\in\mathbb{R}^{16}\times\mathbb{R}$ are generated by a linear model $y=\beta^\top x+e$. The regression coefficient vector $\beta$ for different tasks is randomly sampled with equal probability from two isotropic Gaussian distributions on $\beta$: with means $\beta_0=-\mathbf{4}$ and $\beta_0=\mathbf{4}$, where $\mathbf{4}$ denotes the vector of all fours. The additive noise $e$ is assumed to be white and uncorrelated with $x$, and distributed as the multivariate normal distribution. We consider two cases of $T_{tr}=32$ and $T_{tr}=64$ training tasks, and evaluate the performance of the MAML, Meta-SGD, and TANML on a test set of $T_{v}=64$ tasks. The NMSE performance on the meta-test set obtained by averaging over $30$ Monte Carlo realizations of tasks is reported Table \ref{tab:lr}. 

We observe that both MAML and Meta-SGD perform very poorly in comparison to TANML; Meta-SGD performs slightly better than MAML. Further, we observe that TANML with the Cosine kernel performs the best among the four algorithms. The superior performance of TANML could be ascribed its the nonlinear nature with the gradients enter the estimation through the kernels. The adaptation function involves terms with products of different gradients acting in the spirit of a higher order method unlike MAML/Meta-SGD that use a first order adaptation. This also corroborates with the findings of the recent theoretical work by \cite{saunshi2020sample}, where they show that MAML-type approaches can fail under convex settings (as is the case in this experiment). It is also interesting to note that the value of $\pmb\theta_0$ we obtain for TANML almost coincides with $\mathbf{0}$, which is the value for $\pmb\theta_0$ that theoretically minimizes the average error for this problem. 
We also find that both TANML-Cosine and TANML-Gaussian converge typically in about 5000 iterations, whereas MAML and Meta-SGD do not show improvement in NMSE even after 30000 iterations.}
\begin{table}[t]
\centering
\begin{tabular}{|c|c|c|c|c|}
\hline
    Algorithm & MAML&{Meta-SGD}&TANML-Gaussian&TANML-Cosine\\
    \hline
    $T_{tr}=32$&0.95&0.91&{\bf 0.185}&{\bf 0.079}\\
    $T_{tr}=64$& 0.91&0.86&{\bf 0.15}&{\bf 0.070}\\
    \hline
\end{tabular}
\caption{NMSE on test tasks for the regression experiment 1. }
\label{tab:lr}
\end{table}
\subsection{Experiment 2}
In this experiment, we consider the task of training of non-linear predictors which correspond to a fully connected ANN. 
We consider data pairs $(x,y)\in\mathbb{R}\times\mathbb{R}$ generated from the sinusoidal data source $y=A\sin(\omega x)$, where $x$ is drawn randomly from the interval $[-1,1]$, and $A$ and $\omega$ differ across tasks. We do not use the knowledge that the data comes from a sinusoidal source while training the predictors. We are given $K=4$ shots or data-pairs in each task. 
In order to illustrate the potential of TANML in using the similarity/ dissimilarity among tasks, we consider a fixed fraction of the tasks to be outliers, that is, generated from a non-sinusoidal data source in both meta-training and meta-test data, as described next. The predictor {\color{black}$\hat{y}=f(x,\pmb\theta)$} is the output of a fully-connected four-layer feed-forward neural network of $16$ hidden units in each layer, with Rectified linear unit (ReLU) as the activation function; $\pmb\theta$ is the vector of all the weights and biases in the neural network. 

We consider two different regression experiments:\\
    (1) {\em Experiment 2a $-$ Fixed frequency varying amplitude:}
The data for the different tasks generated from sinusoids with $A$ drawn randomly from $(0,1]$, setting $\omega=1$. The outlier task data generated as $y(x)=Ax$.\\
(2) {\em Experiment 2b $-$ Fixed amplitude varying frequency:}
The data for the different tasks generated from sinusoids with  $\omega$ randomly drawn from $[1,1.5]$, setting $A=1$. The outlier task data is generated as $y(x)=\omega x$.

	 In order that the structural similarities are better expressed, we update kernel regression for the parameters of the different layers separately. {\color{black}This is because using a single adaptation function all components of $\pmb\theta_i$ might result in certain parameters dominating the kernel regression, specially when the dimension of the parameters becomes large. Hence, we perform the adaptation separately for components of $\pmb\theta_i$ corresponding to the different layers $l=1,\cdots,L$ using the adaptation functions $\mathbb{R}^{D_l}\times\mathbb{R}^{D_l\times T_{tr}}\mathbb{R}^{Kn_x}\times \mathbb{R}^{Kn_y}\ni(\pmb\theta_{0,l},\pmb\Psi_l,\mathcal{X},\mathcal{Y})\mapsto g_{\mathrm{TANML},l}(\pmb\theta_{0,l},\pmb\Psi_l,\mathcal{X},\mathcal{Y})\in\mathbb{R}^{D_l}$ for the parameter $\pmb\theta_{i,l}$ belonging to the $l$th network layer}:
$
     \quad\pmb\theta_{i,l}=g_{\mathrm{TANML},l}(\pmb\theta_{0,l},\pmb\Psi_l,\mathcal{X}_i,\mathcal{Y}_i)=\sum_{i'=1}^{{T}_{tr}}\pmb\psi_{i',l}\,k(\mathbf{z}_i(\pmb\theta_{0,B}),\mathbf{z}_{i'}(\pmb\theta_{0,l})),\,\,l=1,\cdots,L.
$
We perform the experiments with the number of meta-training tasks equal to $T_{tr}=256$ and $T_{tr}=512$.
The NMSE performance on test tasks obtained by averaging over $100$ Monte Carlo realizations of tasks is reported Table \label{tab:256}. We observe that TANML outperforms both MAML and Meta-SGD in test prediction by a significant margin even when the fraction of the outlier tasks is $10\%$ and $20\%$. This clearly supports our intuition that an explicit awareness or notion of similarity aids in the learning, specially when the number of training tasks is limited. 
We also observe that on an average TANML with the cosine kernel performs better than the Gaussian kernel. This may perhaps be explained as a result of the Gaussian kernel having an additional hyperparameter that needs to be specified, whereas the cosine kernel does not have any hyperparameters. As a result, the performance of the Gaussian kernel may be sensitive to the choice of the variance hyperparameter $\sigma^2$ and the dataset used. We note that the performance of the approaches in Experiment 1 is better than that in Experiment 2. This is because there is higher variation among the tasks (changing frequency) than in Experiment 1 (changing amplitudes). We also observe that the performance improves slightly as $T_{tr}$ is increased from $256$ to $512$. 
\begin{table}[t]
\centering
\begin{tabular}{|c|c|c|c|c|c|c|c|c|}
\hline
    Algorithm & \multicolumn{2}{c|}{Experiment 2a}& \multicolumn{2}{c|}{Experiment 2a}& \multicolumn{2}{c|}{Experiment 2b}& \multicolumn{2}{c|}{Experiment 2b} \\
    &\multicolumn{2}{c|}{$10\%$ outlier}&\multicolumn{2}{c|}{$20\%$ outlier}&\multicolumn{2}{c|}{$10\%$ outlier}&\multicolumn{2}{c|}{$20\%$ outlier}\\
    \hline
    $T_{tr}$&256&512&256&512&256&512&256&512\\
    \hline
    MAML&0.83&.77&0.75&0.74&0.89&0.81&0.83&0.76\\
    {Meta-SGD}&0.92&1.04&0.81&0.93&1.5&0.92&1.06&0.93\\
         TANML-Gaussian&{\bf0.4}&{\bf 0.41}&{\bf0.38}&{\bf0.38}&{\bf0.76}&{\bf0.60}&{\bf0.73}&{\bf0.58}\\
     TANML-Cosine&{\bf 0.37}&{\bf 0.35}&{\bf 0.30}&{\bf 0.26}&{\bf 0.44}&{\bf 0.38}&{\bf 0.47}&{\bf 0.33}\\
     \hline
\end{tabular}
\caption{NMSE on test tasks for regression experiment 2. }
\label{tab:256}
\vspace{-.25in}
\end{table}
\section{Conclusion}
We proposed a task-similarity aware meta-learning algorithm that explicitly quantifies and employs a similarity between tasks through nonparametric kernel regression. We showed how our approach brings a novel connection between meta-learning and reproducing kernel Hilbert spaces. Our hypothesis was that an explicit incorporation of task-similarity helps improve the meta-learning performance in the task-limited setting with possible outlier tasks. Experiments with multiple regression tasks support our hypothesis, and our algorithm was shown to outperform the popular meta-learning algorithms by a significant margin. The aim of the current contribution was to investigate how task-similarity could be meaningfully employed and used to advantage in meta-learning. To that end, we wish to reiterate that the study is an ongoing one and the experiments considered in this paper are in no way exhaustive. We will be particularly pursuing the application of our approach to classification and few-shot learning tasks in the future. An important next step for our approach is also the use of online/sequential kernel regression techniques to run our algorithm in a sequential or batch-based manner. The nonparametric kernel regression framework also opens doors to a probablistic or Bayesian treatment of meta-learning that we plan to pursue in the recent future. 

\newpage
\bibliography{refs_iclr2021}
\newpage
\appendix

We compare four different approaches: MAML, Meta-SGD, TANML-Cosine, TANML-Gaussian. All the algorithms were trained for 60000 meta-iterations, where each meta-iteration outer update uses the entire set of training tasks, and not as a stochastic gradient descent. All the experiments were performed on either NVIDIA Tesla K80 GPU on Microsoft Azure Platform.

\section{Experiment 2 Hyper-parameters}
\label{app:settings}
The hyper-parameters for the four approaches are listed below. The learning-rate parameters were chosen such that the training error converged without instability.

\subsection{MAML}
\begin{itemize}
    \item Inner update learning rate: $\alpha$: 0.01
    \item Outer update learning rate: $5\times10^{-4}$
    \item Total ANN layers: 4 with, 2 hidden layers
    \item Non-linearity: ReLU
    \item Optimizer: Adam
\end{itemize}
\subsection{Meta-SGD}
\begin{itemize}
    \item Inner update learning rate $\pmb\alpha$: learnt, initialized with values randomly drawn from $[0.001,0.01]$
    \item Outer update learning rate for $\pmb\theta_0$: $5\times10^{-4}$
    \item Outer update learning rate for $\pmb\alpha$: $1\times10^{-6}$
    (Note that the learning rates for $\pmb\theta_0$ and $\pmb\alpha$ are different)
    \item Total ANN layers: 4 with, 2 hidden layers
    \item Non-linearity: ReLU
    \item Optimizer: Adam
\end{itemize}
\subsection{TANML-Gaussian}
\begin{itemize}
    
    \item Outer update learning rate for $\pmb\theta_0$: $1\times10^{-3}$
    \item Outer update learning rate for $\pmb\Psi$: $5\times10^{-5}$
    (Note that the learning rates for $\pmb\theta_0$ and $\pmb\Psi$ are different)
    \item $\mu=0.1$
    \item $\sigma^2=0.5$
    \item Total ANN layers: 4 with, 2 hidden layers
    \item Non-linearity: ReLU
    \item Optimizer: Adam
\end{itemize}

\subsection{TANML-Cosine}
\begin{itemize}
    
    \item Outer update learning rate for $\pmb\theta_0=\pmb\theta_0$: $5\times10^{-4}$
    \item Outer update learning rate for $\pmb\Psi$: $1\times10^{-5}$
     (Note that the learning rates for $\pmb\theta_0$ and $\pmb\Psi$ are different)
    \item $\mu=0.1$
    \item Total ANN layers: 4 with, 2 hidden layers
    \item Non-linearity: ReLU
    \item Optimizer: Adam
\end{itemize}

\section{Experiment 1 hyperparameters}
Except that the predictor was a linear one and there was no ANN involved, that is the parameter was a 16-dimensional vector, all other hyperparamters were kept the same as in Experiment 2. The parameter $\sigma^2$ was set to 10 as it gave the best results.

\end{document}